\documentclass[lettersize,journal]{IEEEtran}
\usepackage{amsmath,amsfonts} 
\usepackage{array}
\usepackage[caption=false,font=normalsize,labelfont=sf,textfont=sf]{subfig}
\usepackage{textcomp}
\usepackage{stfloats}
\usepackage{url}
\usepackage{verbatim}
\usepackage{graphicx}
\usepackage{multirow}
\usepackage{algorithm}
\usepackage{algpseudocode}
\usepackage{booktabs,array,xcolor}
\usepackage{cite}
\begin{document}

\title{LIORNet: Self-Supervised LiDAR Snow Removal Framework \\ for Autonomous Driving under Adverse Weather Conditions}

\author{Ji-il Park, Inwook Shim* 
\thanks{Ji-il Park is with the Ministry of National Defense, 22, Itaewon-ro, Yongsan-gu, Seoul, Republic of Korea. (e-mail:\tt\footnotesize tinz64@kaist.ac.kr)}
\thanks{Inwook Shim is with the Department of Smart Mobility Engineering, Inha University, Republic of Korea.~({e-mail:\tt\footnotesize iwshim@inha.ac.kr})}}




\maketitle

\begin{abstract}
LiDAR sensors provide high-resolution 3D perception and long-range detection, making them indispensable for autonomous driving and robotics. However, their performance significantly degrades under adverse weather conditions such as snow, rain, and fog, where spurious noise points dominate the point cloud and lead to false perception. To address this problem, various approaches have been proposed: distance-based filters exploiting spatial sparsity, intensity-based filters leveraging reflectance distributions, and learning-based methods that adapt to complex environments. Nevertheless, distance-based methods struggle to distinguish valid object points from noise, intensity-based methods often rely on fixed thresholds that lack adaptability to changing conditions, and learning-based methods suffer from the high cost of annotation, limited generalization, and computational overhead. In this study, we propose LIORNet, which eliminates these drawbacks and integrates the strengths of all three paradigms. LIORNet is built upon a U-Net++ backbone and employs a self-supervised learning strategy guided by pseudo-labels generated from multiple physical and statistical cues, including range-dependent intensity thresholds, snow reflectivity, point sparsity, and sensing range constraints. This design enables LIORNet to distinguish noise points from environmental structures without requiring manual annotations, thereby overcoming the difficulty of snow labeling and the limitations of single-principle approaches. Extensive experiments on the WADS and CADC datasets demonstrate that LIORNet outperforms state-of-the-art filtering algorithms in both accuracy and runtime while preserving critical environmental features. These results highlight LIORNet as a practical and robust solution for LiDAR perception in extreme weather, with strong potential for real-time deployment in autonomous driving systems.
\end{abstract}


\section{Introduction} 
\IEEEPARstart{L}{iDAR} has become a core perception sensor in autonomous driving and robotics thanks to its ability to capture dense 3D information with high resolution, wide field-of-view, and robustness to illumination changes. Beyond vehicles, LiDAR is now widely adopted in drones, unmanned ground vehicles (UGVs), unmanned surface vehicles (USVs), and surveillance systems, underscoring its growing importance in enabling reliable perception across diverse applications. Despite these advantages, LiDAR remains highly vulnerable to adverse weather such as snow, rain, and fog. In such conditions, atmospheric scattering and multiple reflections generate a large number of spurious points, which contaminate the point cloud and significantly degrade downstream perception tasks such as detection, tracking, and localization.

To address this challenge, three main categories of LiDAR denoising approaches have been investigated. Distance-based methods leverage spatial sparsity or range consistency to filter outliers, offering computational efficiency but often removing valid object points, thereby reducing accuracy. Intensity-based methods exploit the reflectivity properties of snow particles to achieve efficient filtering. The LIOR algorithm, for example, applies a fixed intensity threshold to suppress noise points with high accuracy and speed, but suffers from limited adaptability to environmental changes. To overcome this limitation, D-LIOR introduced a dynamic thresholding strategy, estimating snow intensity adaptively via regression, and further improved efficiency through reclassification and parallelization. While these approaches demonstrated strong performance, they still rely on handcrafted physical criteria and may struggle to generalize across varying conditions. Learning-based methods, on the other hand, employ deep neural networks to capture complex noise distributions, showing strong adaptability and representation power. However, they require large-scale annotated datasets, face generalization challenges across different sensors and weather conditions, and impose high computational costs, which hinder real-time deployment.

These limitations highlight the need for a new framework that integrates the strengths of distance-, intensity-, and learning-based paradigms while mitigating their drawbacks. In this paper, we propose LIORNet, a hybrid LiDAR snow-removal network designed for robust perception in extreme weather. LIORNet is built upon a U-Net++ backbone and adopts a self-supervised learning strategy guided by pseudo-labels automatically generated from multiple physical and statistical cues, including range-dependent intensity thresholds, snow reflectivity, point sparsity, and sensing range constraints. This design enables LIORNet to suppress weather-induced noise while preserving environmental structures without requiring manual annotations, thus overcoming the difficulty of snow labeling and the limitations of single-principle approaches. 

The key contributions of this work are as follows: 
\begin{itemize}
    \item \textbf{First approach unifying existing methods:} A LiDAR snow-removal framework that integrates distance-based, intensity-based, and learning-based approaches.
    \item \textbf{Physics-based pseudo-label generation:} Pseudo-labels designed by incorporating the physical properties of snow and sensor constraints.
    \item \textbf{Self-supervised learning without labeled data:} Snow mask generation guided by pseudo-labels using a U-Net++ architecture.
    \item \textbf{Implementation of U-Net++ for snow masking:} U-Net++, widely used for segmentation tasks, is applied for the first time to LiDAR snow removal, demonstrating its effectiveness in self-supervised snow masking.
    \item \textbf{Validation under diverse snowfall conditions:} High accuracy and efficiency demonstrated on datasets such as WADS and CADC.
    \item \textbf{Scalability to annotation-scarce environments:} Applicability confirmed in real-world snowy conditions where labeling is impractical.
\end{itemize}

\section{Literature Review}
\subsection{Distance-based methods}
Distance-based filters exploit spatial sparsity and range consistency to suppress outliers in LiDAR point clouds. Early statistical variants such as Radius Outlier Removal~(ROR) and Statistical Outlier Removal~(SOR) were not designed for adverse-weather artifacts and thus suffered from limited accuracy and efficiency. Dynamic Radius Outlier Removal~(DROR) tackles this by expanding the search radius with increasing range, yielding markedly higher accuracy in snowfall than ROR albeit at significant computational cost~\cite{charron2018noising}. Fast Cluster Statistical Outlier Removal~(FCSOR) improves throughput on large outdoor point clouds via voxel subsampling and partial parallelization, yet it does not specifically target weather noise~\cite{balta2018fast}. Dynamic Statistical Outlier Removal~(DSOR) introduces KNN-based distance statistics to derive a dynamic threshold, stabilizing performance compared with DROR~\cite{kurup2021dsor}. PCA-based methods such as PCA-based Adaptive Clustering~(PCAAC) and PCA-based Adaptive Radius Outlier Removal~(PCAAR) reduce dimensionality and adopt adaptive clustering and radius control to improve both accuracy and runtime~\cite{duan2021low1,duan2021low2}. More recently, Adaptive Outlier Removal filter on range Image~(AORI) projects $3$D point clouds to $2$D range images to cut computation~\cite{le2023efficient}, while Dynamic Channel-Wise Outlier Removal~(DCOR) applies channel-wise filtering with dynamic radius adaptation to boost both precision and speed under snow~\cite{zhou2024dcor}. Despite these advances, distance-based filters remain computation-intensive and risk removing valid object points at long ranges.

\begin{figure*}[t!]
\centerline{\includegraphics[width=0.635\textwidth]{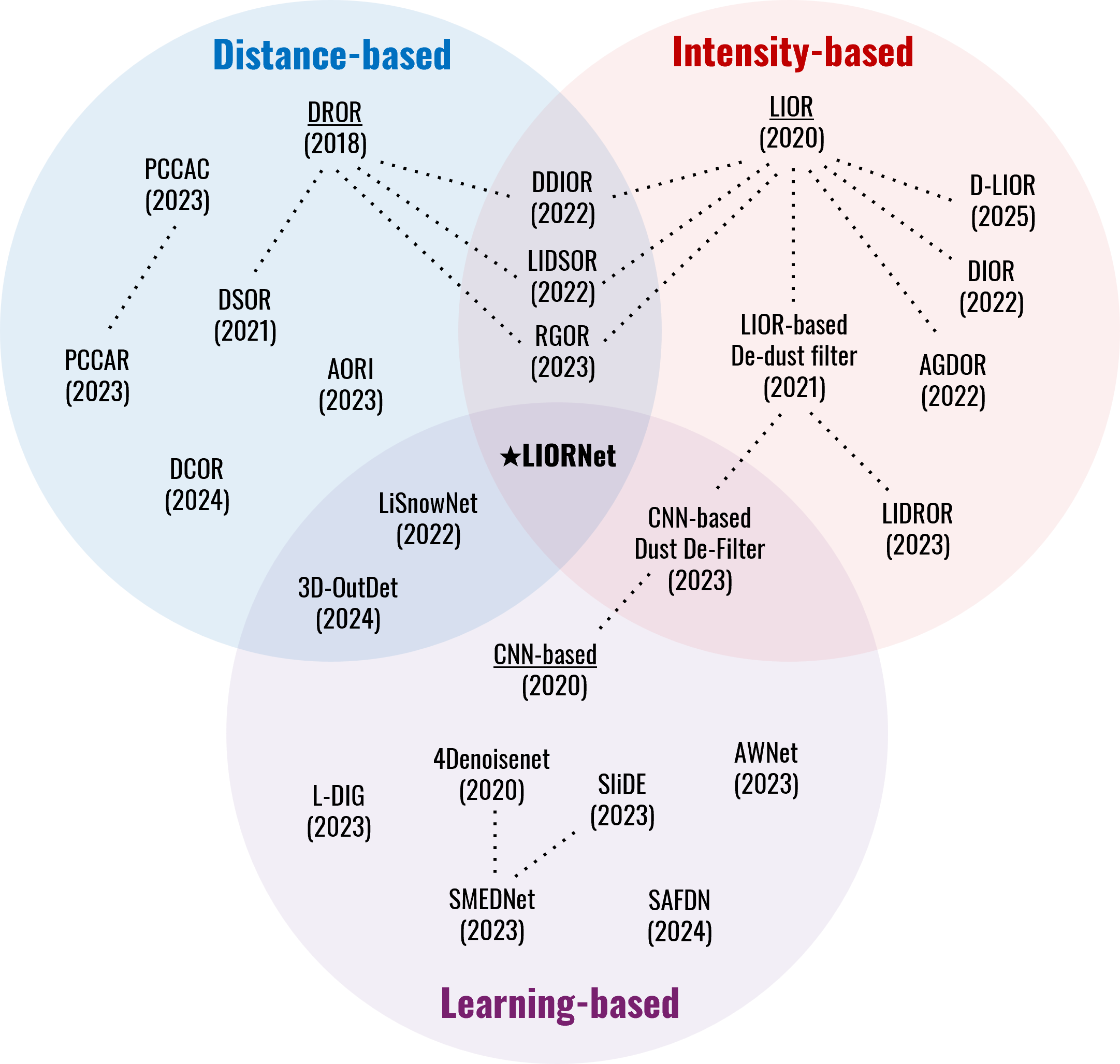}}
\caption{Categorization by approach \cite{park2025review}. LIORNet shown at the center is the first attempt to unify all three strategies.}
\label{fig:approach}
\end{figure*}

\subsection{Intensity-based methods}
Intensity-based approaches leverage reflectance characteristics of atmospheric particles such as snow, rain, and dust. Low Intensity Outlier Removal~(LIOR) pioneered this line by deriving a physically motivated fixed intensity threshold and restoring misclassified points through a Radius Inlier Saving~(RIS) step, outperforming DROR in both speed and accuracy~\cite{park2020fast}. Extensions tailored to dust include a LIOR-based de-dust filter and Low Intensity Dynamic Radius Outlier Removal~(LIDROR), which pair fixed thresholds with ROR and DROR restoration~\cite{afzalaghaeinaeini2021design, afzalaghaeinaeini2022design}. Dynamic Intensity Outlier Removal~(DIOR) combines LIOR and DROR with Field Programmable Gate Array~(FPGA) acceleration to achieve real-time performance, but still relies on a fixed threshold on the intensity stage~\cite{roriz2021dior}. To increase adaptability, the Adaptive Two-Stage filter bins intensity into low, medium, and high ranges prior to distance-based search~\cite{le2022adaptive1}; Adaptive Group of Density Outlier Removal~(AGDOR) adopts an adaptive density strategy on WADS~\cite{le2022adaptive2}; a neighbor-intensity method averages KNN reflectance to better separate snow from structure~\cite{kwon2023neighbor}; and a rain-specific method first restores raindrop-induced intensity distortions before suppressing low-intensity noise~\cite{han2023denoising}. Dynamic Low Intensity Outlier Removal~(D-LIOR) extends LIOR by introducing an adaptive intensity threshold that improves robustness across varying snow densities~\cite{dlior2025}. While this enhancement increases adaptability compared with fixed-threshold methods, its reliance on manually designed rules remains a limiting factor. Although intensity-based filters are generally faster and more accurate than distance-only methods, their rule-driven nature reduces robustness across different sensors and environments.

\subsection{Distance–Intensity fusion methods}
Fusion methods combine range and reflectance cues to overcome the shortcomings of single-feature filtering. Dynamic Distance-Intensity Outlier Removal (DDIOR) jointly thresholds distance and intensity, effectively extending DSOR and surpassing DROR and DSOR on WADS while preserving fine structures \cite{wang2022scalable}. Low Intensity Dynamic Statistical Outlier Removal (LIDSOR) fuses LIOR and DSOR but adapts detection distances via gamma-distribution fitting to accommodate snow and rain variability \cite{huang2023lidsor}. A temporal filtering method integrates distance–intensity cues with frame-to-frame coherence, using Time Outlier Removal (TOR) and Entropy Weight Method (EWM) to distinguish snow from object motion and suppress both noise and ghosting \cite{yan2024denoising}. Reflectance and Geometrical Outlier Removal (RGOR) restores reflectance, groups beams geometrically, and applies block-wise filtering followed by geometry-aware refinement, reaching performance comparable to learning-based LiSnowNet and 4DenoiseNet on WADS \cite{han2023rgor}. Conditional Random Field Outlier Removal (CRFOR) formulates denoising as a spatio-temporal CRF, propagating labels from high-confidence nodes using physical features such as intensity and distribution, outperforming DROR, LIOR, and DDIOR while rivaling learning-based methods \cite{wang2023snow}. Beyond range–intensity, density–intensity fusion that first isolates snowfall regions via density and then removes noise with intensity fusion has also shown gains over DSOR and DDIOR \cite{pan2023lidar}. Overall, fusion-based methods improve both accuracy and speed by leveraging the complementary strengths of distance and intensity features. However, they often involve complex algorithmic structures and still inherit certain limitations of conventional filters, such as threshold dependence.

\subsection{Learning-based methods}
Learning-based approaches directly model complex noise distributions from data. The first CNN-based model WeatherNet was trained with synthetic fog and rain augmentations and outperformed DROR and RangeNet, demonstrating the promise of data-driven filtering \cite{heinzler2020cnn}. 4DenoiseNet leverages spatio-temporal coherence across adjacent frames to better preserve dynamic objects and remove weather noise \cite{seppanen20224denoisenet}. LiSnowNet adapts Multi-Level Wavelet Convolutional Neural Network (MWCNN) \cite{liu2019multi} with residual blocks and dropout for snow, achieving strong results on the Canadian Adverse Driving Conditions (CADC) dataset \cite{pitropov2021canadian} and the Winter Adverse Driving Scenarios (WADS) dataset \cite{kurup2021dsor}, as well as in follow-up benchmarks \cite{yu2022lisnownet}. Self-supervised learning has emerged to mitigate annotation costs: Self-supervised LiDAR De-snowing through Reconstruction Difficulty (SLiDE) uses reconstruction difficulty as a training signal \cite{bae2022slide}, and multi-echo self-supervised pipelines improve denoising without dense labels \cite{seppanen2023multi}. Generative paradigms extend capability further: LiDAR depth images GAN (L-DIG) employs GANs to both remove and synthesize snow \cite{zhang2023dig}, while GAN inversion has been explored for fog restoration \cite{chai2023gan}. For efficiency, 3D-OutDet introduces neighborhood convolutions to reduce memory and achieve state-of-the-art end-to-end speed \cite{raisuddin20243d}. Energy-based outlier detection has also been investigated under adverse weather \cite{piroli2023energy}. TripleMixer combines geometric, frequency, and channel cues to achieve state-of-the-art results across snow, rain, and fog \cite{zhao2024triplemixer}. Nonetheless, learning-based methods still face costly annotation, cross-sensor and domain generalization, and compute constraints that challenge real-time deployment. In particular, while CNN, range image, and GNN-based approaches have substantially improved accuracy, they remain slower than rule-based filters, limiting their applicability in real-time scenarios. Moreover, the scarcity of point-wise labeled datasets in extreme environments significantly hinders scalability, underscoring dataset construction as a critical direction for future research.

\subsection{Limitation}
Overall, existing approaches present complementary strengths but also notable weaknesses. Distance-based methods have improved accuracy but remain computationally intensive and prone to discarding valid object points at longer ranges. Intensity-based methods achieve faster and more accurate filtering than distance-only schemes, yet their reliance on sensor- and environment-dependent thresholds limits generalization. Fusion-based methods combine the advantages of distance and intensity cues, enhancing both accuracy and runtime, but their algorithmic complexity and partial inheritance of threshold dependence restrict scalability. Learning-based methods significantly boost accuracy through CNN-based, range image-based, and GNN-based models, but they lag behind rule-based filters in processing speed and are hindered by the scarcity of point-wise labeled datasets in extreme environments. These persistent limitations underscore the need for a hybrid, self-supervised framework that integrates physical and statistical priors with adaptive learning—a direction pursued in LIORNet.

\section{LIORNet : Proposed Framework}

\begin{figure*}[t!]
\centerline{\includegraphics[width=1.0\textwidth]{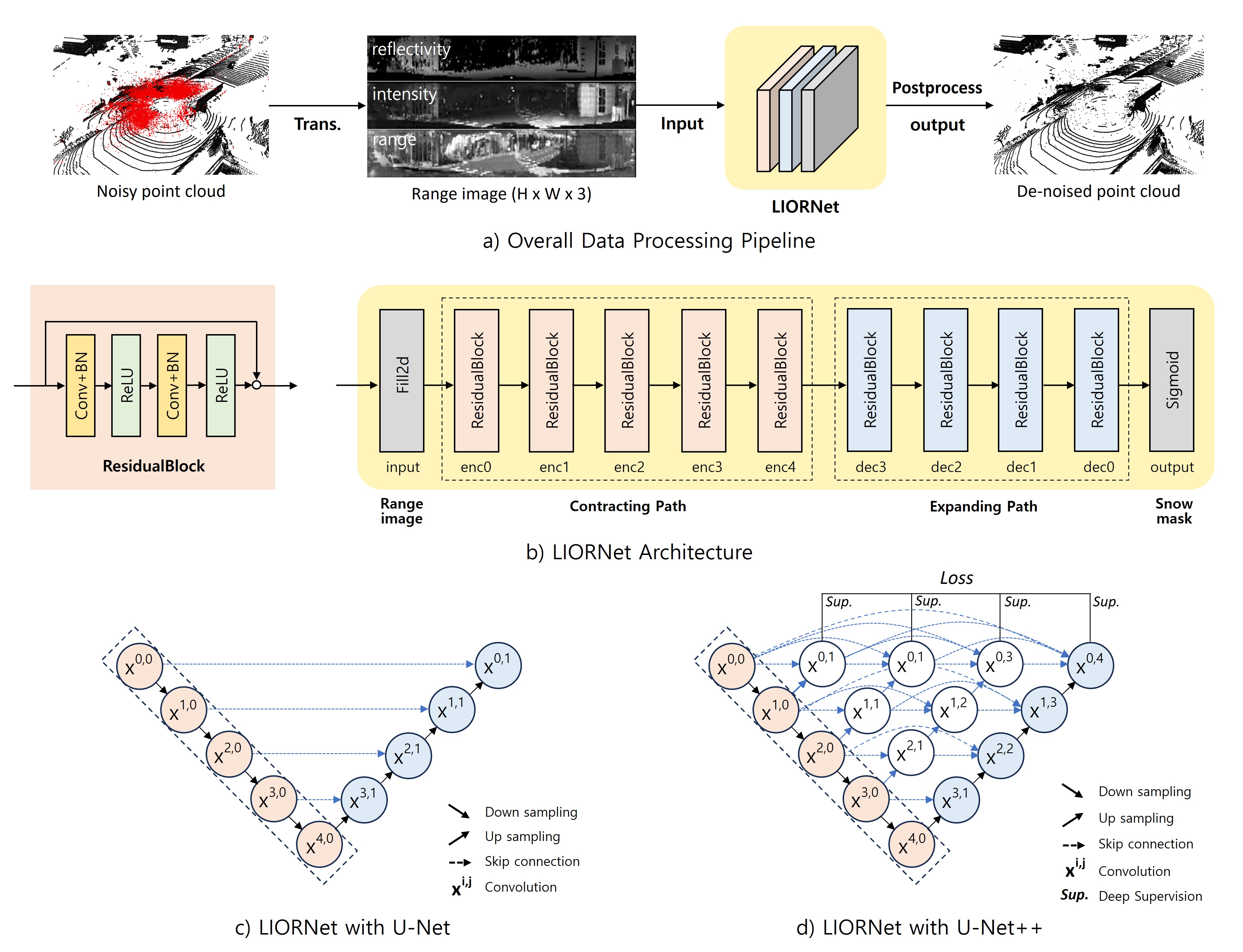}}
\caption{Overall LIORNet framework: (a) data processing pipeline, (b) network architecture, and backbone configurations using (c) U-Net and (d) U-Net++.}
\label{fig:architecture}
\end{figure*}

\subsection{Overview}
We introduce LIORNet, a self-supervised framework leveraging pseudo-labeling to effectively remove snow noise while maintaining critical environmental structures. Studies surveying LiDAR denoising methods in adverse environments report that no algorithm has yet been documented to simultaneously incorporate distance-based, intensity-based, and learning-based approaches \cite{park2025review}. As illustrated in Fig. \ref{fig:approach}, the proposed LIORNet represents the first framework that unifies all three strategies for LiDAR denoising under snowy weather conditions. The core of LIORNet lies in a U-Net++ encoder–decoder architecture optimized for processing LiDAR range–intensity–reflectivity data. The network is trained to detect and suppress snow-induced outliers without relying on explicit ground-truth annotations, thereby mitigating the scarcity of labeled data under adverse weather conditions. To accomplish this, the framework employs pseudo-label generation based on adaptive intensity modeling, reflectivity consistency, and spatial sparsity priors. Furthermore, physics-based penalty terms derived from intensity, reflectivity, and range distributions are incorporated during training to prevent the restoration of unnecessary noise and the removal of valid structural information. Finally, a post-processing step that considers the sensing range characteristics of snow and ground conditions is applied to correct subtle errors that the network may miss, thereby improving overall accuracy. In this way, LIORNet establishes itself as a unified framework encompassing rule-based filtering, self-supervised learning, physics-based penalties, and post-processing, ensuring high generalization across diverse LiDAR sensors and snowy environments.

\begin{figure*}[t!]
\centerline{\includegraphics[width=0.95\textwidth]{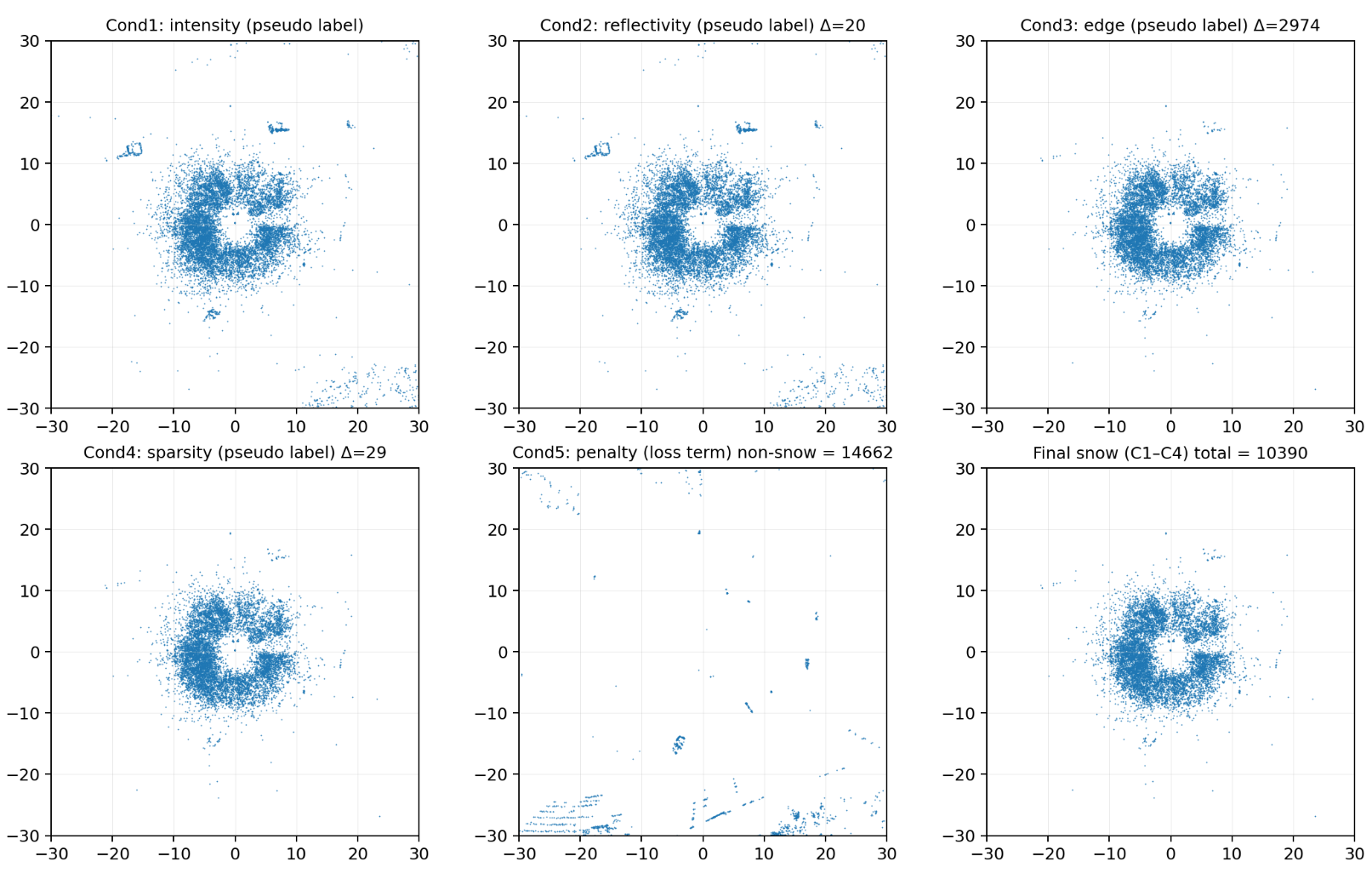}}
\caption{Visualization of conditional pseudo-label point updates. Candidate snow points are sequentially refined through intensity (Cond1), reflectivity (Cond2), edge (Cond3), and sparsity (Cond4) conditions resulting in the final pseudo-labeled snow points.}
\label{fig:pseudo}
\end{figure*}

\subsection{Pipeline and Architecture}
The overall pipeline of LIORNet is shown in Fig.~\ref{fig:architecture}. Noisy point clouds acquired under snowy weather are first transformed into range images with three channels: range, intensity, and reflectivity. This transformation provides a structured 2D representation of the 3D LiDAR measurements, facilitating convolution-based processing. The resulting range images are then fed into the training pipeline for model learning, and during the inference stage the final de-noised point clouds are produced.

At the network level, LIORNet adopts a U-Net++ encoder--decoder backbone composed of residual blocks, each consisting of convolution, batch normalization, and ReLU layers. Residual connections promote feature reuse and stabilize training under sparse and noisy conditions. The contracting path extracts hierarchical features at progressively coarser scales through $2\times$ pooling, while the expanding path reconstructs refined representations via bilinear upsampling and convolutional refinement. Four skip connections link encoder and decoder features to preserve spatial details and improve localization accuracy. 

Intermediate decoder outputs are further refined through residual blocks, and the U-Net++ principles of nested connections and deep supervision are employed to enhance performance. Fig.~\ref{fig:architecture} (c) illustrates LIORNet with a basic U-Net structure, while (d) shows LIORNet with a U-Net++ structure incorporating deep supervision. Their predictions are upsampled to the full resolution and jointly optimized with the main output, thereby facilitating multi-scale feature learning.

The final prediction head consists of a convolution layer followed by a sigmoid activation, producing a dense binary snow mask aligned with the range image. This mask represents the likelihood of each measurement being snow-induced noise and is applied to the original point cloud on a one-to-one basis to generate the de-noised output. Through this design, LIORNet combines a systematic data processing pipeline with a U-Net++-based architecture, enabling robust snow noise removal while preserving essential structural details of the environment.

\subsection{Self-Supervision Strategy}
Point-wise labeling under adverse snowy conditions is highly labor-intensive and time-consuming, making it impractical for large-scale LiDAR datasets. To address this challenge, LIORNet adopts a self-supervised learning strategy that automatically generates pseudo-labels by combining physics-based constraints with empirical rules derived from data. 

Specifically, LIORNet constructs pseudo-labels based on four complementary conditions. 1) Intensity-guided candidate selection: Snow candidates are first extracted by applying a physically derived intensity threshold, after correcting for the effective sensing range of snow and ground geometry. 2) Reflectivity-based snow discrimination: Among these candidates, points with near-zero reflectivity are retained as snow, while others are discarded. 3) Edge-aware restoration: Points overlapping with structural edges detected from the range map are excluded to avoid misclassifying boundaries as snow. 4) Density-based object removal: Dense clusters exceeding distance-dependent occupancy thresholds are regarded as real objects, and candidate snow points in these regions are removed. The detailed procedure is described in Algorithm 1, and the sequential refinement of pseudo-labels is visualized in Fig.~\ref{fig:pseudo}.

The pseudo-labels generated by these conditions serve as supervisory signals during training. 
To further improve robustness, an uncertainty-based weighting scheme adaptively balances multiple loss components, including intensity, reflectivity, sparsity, and edge. 
This design prevents the network from overfitting to noisy pseudo-labels and stabilizes optimization. As a result, LIORNet can effectively learn to remove snow-induced noise while preserving the physical constraints of the sensor and the geometric structure of the scene, all without the need for manual annotation.

\begin{algorithm}[t!]
\caption{Pseudo-label generation}
\small
\renewcommand{\baselinestretch}{1.025}\selectfont
\begin{algorithmic}   

\For{$p \in P$}
    \Statex // Preprocessing
    \State remove $p$ if $\operatorname{range}(p) > 71.235\,\text{m}$
    \State remove $p$ if $\operatorname{height}(p) \le \text{LiDAR}_H$

    \Statex // Condition 1 : Intensity-guided candidate selection
    \If{$I_p \le I_{\mathrm{thr}}$}
        \State classify $p$ as snow $p_s$
    \Else
        \State classify $p$ as non-snow $p_o$
    \EndIf

    \Statex // Condition 2 : Reflectivity-based snow discrimination
    \If{$R_p \neq 0$}
        \State reclassify $p$ as object $p_o$
    \EndIf

    \Statex // Condition 3 : Edge-aware restoration
    \State apply edge detection on $p$
    \If{$p$ \text{ is an edge point}}
        \State reclassify $p$ as object $p_o$
    \EndIf

    \Statex // Condition 4 : Density-based object removal
    \State count neighbors $n_p$ of $p$ within radius $r$
    \If{$n_p \ge n_{\mathrm{thr}}$}
        \State reclassify $p$ as object $p_o$
    \EndIf
\EndFor

\Statex // Outputs
\State Final snow set $P_s = \{p \mid p \text{ not reclassified to } p_o\}$%

\end{algorithmic}    
\end{algorithm}

\subsection{Loss Function}
The proposed LIORNet is trained in a self-supervised manner using pseudo-labels, where each condition corresponds to a tailored loss term. The overall objective function is defined by five loss terms: 1) an \textit{intensity consistency loss} corresponding to intensity-guided candidate selection, 2) a \textit{reflectivity discrimination loss} corresponding to reflectivity-based snow discrimination, 3) an \textit{edge suppression loss} corresponding to edge-aware restoration, 4) a \textit{Density-aware sparsity loss} corresponding to density-based object removal, and 5) a \textit{threshold penalty loss} that penalizes points violating conditions. Furthermore, an uncertainty-weighted formulation is incorporated, allowing the loss terms to be selectively activated. In practice, the predicted mask is compared against the final pseudo-label refined through Conditions 1–4, and each loss term reflects the rule of its corresponding condition. Detailed descriptions of these loss terms are provided in the following subsections.

\subsubsection*{1) \textit{Intensity consistency loss}}
Candidate snow points are initially extracted by applying a range-dependent intensity threshold to form the initial mask. After sequential refinement through Conditions 2–4, the final pseudo-label is obtained. During training, the predicted mask $\hat{M}$ is compared against the pseudo-label mask $M^{\text{final}}$ using binary cross-entropy~(BCE):
\begin{equation}
\mathcal{L}_{\text{intensity}} = \text{BCE}\left( \hat{M}, M^{\text{final}} \right).
\label{eq:loss_intensity}
\end{equation}

\subsubsection*{2) \textit{Reflectivity discrimination loss}}
The reflectivity loss term~Eq.~\ref{eq:reflectivity} corresponds to pseudo-label Condition 2, training the model to learn the rule that points with zero reflectivity are treated as snow, while points with non-zero reflectivity are classified as objects:
\begin{equation}
\mathcal{L}_{\text{reflectivity}} = \text{BCE}\left( \hat{M}, M^{\text{final}} \right).
\label{eq:reflectivity}
\end{equation}

\subsubsection*{3) \textit{Edge suppression loss}}
The edge suppression loss term~Eq.~\ref{eq:loss_edge} corresponds to pseudo-label Condition~3, training the model to learn the rule that structural boundary pixels detected in the range map should not be classified as snow. Here, $\hat{M}\big|_E$ denotes the prediction restricted to the edge regions $E$, and the loss is computed using the standard binary cross-entropy with logits (\textit{$BCE\_logit$}), which directly takes the raw logits rather than sigmoid probabilities to ensure numerical stability:
\begin{equation}
\mathcal{L}_{\text{edge}} = \text{BCE\_logit}\left( \hat{M}\big|_E, 0 \right).
\label{eq:loss_edge}
\end{equation}

\subsubsection*{4) \textit{Density-aware sparsity loss}}
The density-aware sparsity loss term~Eq.~\ref{eq:loss_density} corresponds to pseudo-label Condition~4, training the model to learn the rule that densely clustered points are more likely to be real objects rather than snow, and thus discouraging overly dense snow predictions. Here, $p_i$ denotes a candidate snow point, $\mathrm{dist}_{k}(p_i)$ represents the mean 
distance to its $k$ nearest neighbors, $w_i = \hat{M}(p_i)$ is the predicted snow probability for point $p_i$ with $w_i \in [0,1]$, and $N$ is the total number of candidate points. To prevent the network from converging to a trivial solution that suppresses all predictions to minimize the sparsity loss, a stop-gradient strategy is applied to this term. Specifically, $\hat{M}(p_i)$ is detached from the gradient flow so that the sparsity loss does not directly influence gradient updates but only functions as a statistical prior:
\begin{equation}
\mathcal{L}_{\text{sparsity}} =
\frac{1}{N} \sum_{i=1}^{N} w_i \cdot \frac{1}{\mathrm{dist}_{k}(p_i)}.
\label{eq:loss_density}
\end{equation}

\subsubsection*{5) \textit{Threshold penalty loss}}
An additional penalty term~Eq.~\ref{eq:loss_penalty} is introduced to complement the existing condition-based losses, as rule-violating predictions may still occur. Specifically, predicted snow points are penalized when they (i) exceed the theoretical intensity threshold, (ii) fall beyond the maximum LiDAR range $d_{\max}$, or (iii) lie below the ground surface. For numerical stability, these penalties are applied in a smoothed form using sigmoid functions. Here, $\Delta I$ denotes the difference between the point intensity and the theoretical threshold intensity, $\Delta d$ represents the range exceeding the maximum LiDAR limit $d_{\max}$, and $\Delta z$ is the vertical offset of a point lying below the ground surface.
\begin{equation}
\mathcal{L}_{\text{thr}} = \alpha \cdot \sigma(\Delta I) + 
                           \beta \cdot \sigma(\Delta d) + 
                           \gamma \cdot \sigma(\Delta z).
\label{eq:loss_penalty}
\end{equation}

\subsubsection*{6) Total loss with uncertainty-based weighting}
The overall training objective is defined by integrating the five condition-based loss terms~Eq.~\ref{eq:loss_total}. 
To balance their relative contributions, an uncertainty-based weighting scheme~\cite{Kendall_2018_CVPR} 
is optionally applied, in which each loss term is adaptively scaled according to its learnable 
log-variance parameter $\log\sigma_i$. Here, $\mathcal{L}_i$ denotes the $i$-th loss term, $\sigma_i$ 
is the learnable uncertainty parameter associated with that loss, and $\lambda_i$ is an ablation 
weight that allows each term to be selectively included or excluded during training. The final objective is defined as follows:
\begin{equation}
\mathcal{L}_{\text{total}} =
\sum_{i=1}^{5} \lambda_i \left(
\frac{\mathcal{L}_i}{2 \exp(\log\sigma_i)} + \frac{\log\sigma_i}{2}
\right).
\label{eq:loss_total}
\end{equation}

\begin{table*}[t!]
\centering
\caption{Quantitative Evaluation of LIORNet under Different Configurations. \\ Postprocess was applied with the theoretical snow sensing limit of 71.23m. \\ (UW: Uncertainty-based Weighting, PP: Postprocessing, DS: Deep Supervision)} 
\renewcommand{\arraystretch}{1.35}   
\setlength{\tabcolsep}{12.05pt}      
\footnotesize                        
\begin{tabular}{|c|c|c|c|c|c|c|c|c|}
\hline
\textbf{ID} & \textbf{UW} & \textbf{PP} & \textbf{DS} &
\textbf{Precision} & \textbf{Recall (TPR)} & \textbf{FPR} & \textbf{FNR} & \textbf{IoU} \\
\hline
1 &  X    & X & X & 0.379 $\pm$ 0.107 & 0.942 $\pm$ 0.088 & 0.265 $\pm$ 0.072 & 0.058 $\pm$ 0.088 & 0.368 $\pm$ 0.104 \\ 
2 &  X    & O & X & 0.473 $\pm$ 0.136 & \textbf{0.961 $\pm$ 0.068} & 0.175 $\pm$ 0.049 & \textbf{0.039 $\pm$ 0.068} & 0.462 $\pm$ 0.132 \\ 
3 &  X    & X & O & 0.402 $\pm$ 0.106 & \textcolor{red}{\textbf{0.939 $\pm$ 0.093}} & 0.240 $\pm$ 0.069 & \textcolor{red}{\textbf{0.061 $\pm$ 0.093}} & 0.389 $\pm$ 0.103 \\ 
\textcolor{blue}{\textbf{4}} &  \textcolor{blue}{\textbf{X}}    & \textcolor{blue}{\textbf{O}} & \textcolor{blue}{\textbf{O}} & \textcolor{blue}{\textbf{0.498 $\pm$ 0.132}} & 0.958 $\pm$ 0.072 & \textcolor{blue}{\textbf{0.157 $\pm$ 0.046}} & 0.042 $\pm$ 0.072 & \textcolor{blue}{\textbf{0.486 $\pm$ 0.128}} \\
\hline
5 & 20\% & X & X & 0.378 $\pm$ 0.104 & 0.941 $\pm$ 0.090 & 0.266 $\pm$ 0.074 & 0.059 $\pm$ 0.090 & 0.367 $\pm$ 0.101 \\ 
6 & 20\% & O & X & \textbf{0.477 $\pm$ 0.133} & 0.960 $\pm$ 0.070 & \textbf{0.172 $\pm$ 0.048} & 0.040 $\pm$ 0.070 & \textbf{0.465 $\pm$ 0.129} \\ 
\textcolor{red}{\textbf{7}} & \textcolor{red}{\textbf{20\%}} & \textcolor{red}{\textbf{X}} & \textcolor{red}{\textbf{O}} & \textcolor{red}{\textbf{0.337 $\pm$ 0.095}} & 0.945 $\pm$ 0.084 & \textcolor{red}{\textbf{0.319 $\pm$ 0.084}} & 0.055 $\pm$ 0.084 & \textcolor{red}{\textbf{0.329 $\pm$ 0.092}} \\ 
8 & 20\% & O & O & 0.444 $\pm$ 0.124 & \textbf{0.962 $\pm$ 0.066} & 0.196 $\pm$ 0.048 & \textbf{0.038 $\pm$ 0.066} & 0.435 $\pm$ 0.120 \\
\hline
9 & 100\%& X & X & 0.351 $\pm$ 0.104 & 0.945 $\pm$ 0.085 & 0.302 $\pm$ 0.079 & 0.055 $\pm$ 0.085 & 0.342 $\pm$ 0.102 \\ 
10 & 100\%& O & X & 0.442 $\pm$ 0.135 & \textcolor{blue}{\textbf{0.964 $\pm$ 0.065}} & 0.200 $\pm$ 0.053 & \textcolor{blue}{\textbf{0.037 $\pm$ 0.065}} & 0.433 $\pm$ 0.132 \\ 
11 & 100\%& X & O & 0.394 $\pm$ 0.107 & 0.940 $\pm$ 0.091 & 0.248 $\pm$ 0.069 & 0.060 $\pm$ 0.091 & 0.382 $\pm$ 0.104 \\ 
\textbf{12} & \textbf{100\%} & \textbf{O} & \textbf{O} & \textbf{0.488 $\pm$ 0.134} & 0.959 $\pm$ 0.071 & \textbf{0.164 $\pm$ 0.047} & 0.041 $\pm$ 0.071 & \textbf{0.476 $\pm$ 0.130} \\
\hline
\end{tabular} 
\label{tab:model_compare}
\end{table*}

\section{Experimental Results}
\subsection{Configuration Options}
\begin{figure*}[t!]
\centerline{\includegraphics[width=0.95\textwidth]{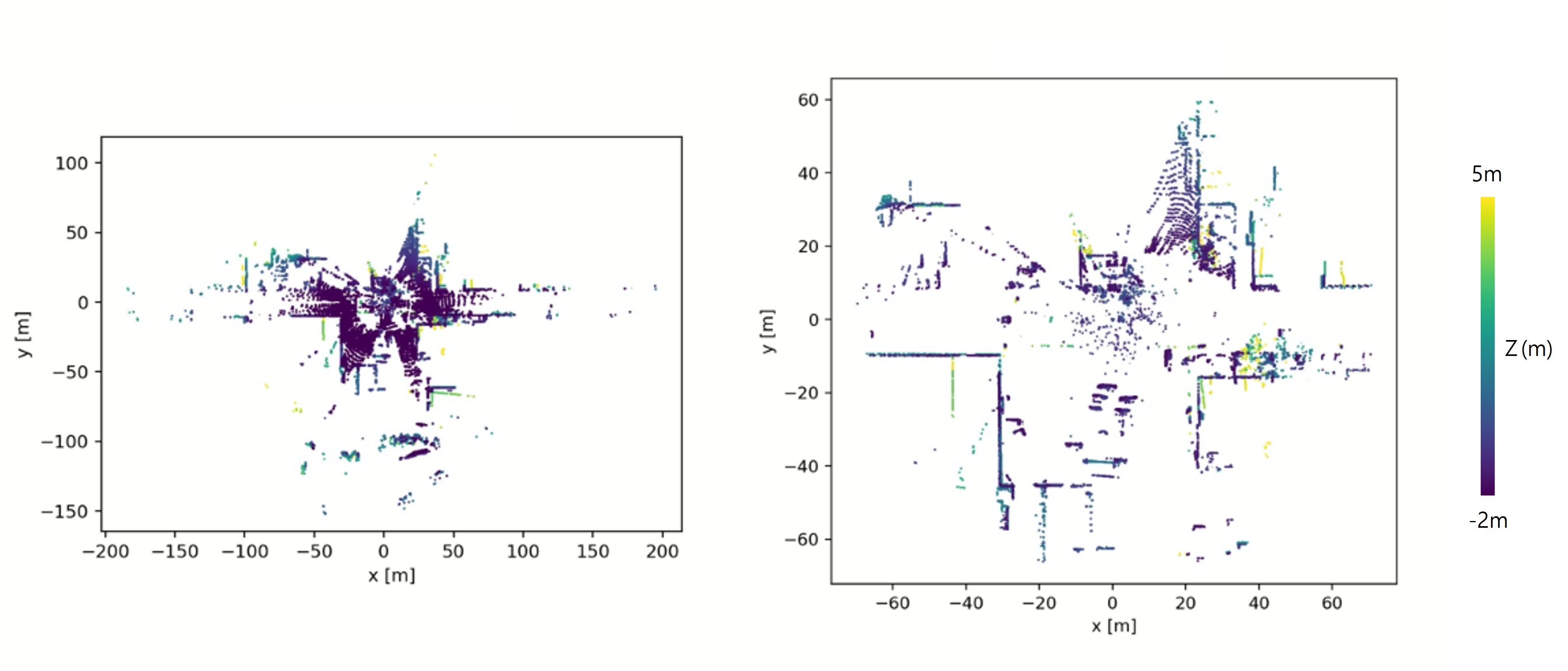}}
\caption{False-positive snow data before and after applying the rule-based postprocessing (considering LiDAR height and the theoretical snow-sensing distance).}
\label{fig:FP_compare}
\end{figure*}

LIORNet incorporates three configuration options: uncertainty-based weighting (UW), postprocessing (PP), and deep supervision (DS). UW dynamically adjusts the relative contribution of multiple loss components during training, thereby stabilizing optimization. DS introduces auxiliary outputs at reduced resolutions to improve feature learning across multiple scales. PP, although rule-based, is a powerful step that restores snow points located beyond the theoretical sensing distance of 71.235 m, which is determined by LiDAR mounting height and ground geometry. As illustrated in Fig.~\ref{fig:FP_compare}, PP effectively suppresses false positives and enhances scene integrity.

The comparative evaluation of different configurations is summarized in Table~\ref{tab:model_compare}. Among the tested conditions, the fourth case—where uncertainty-based weighting is disabled (fixed weights are used), but both PP and DS are applied—achieves the most competitive results. In contrast, the lowest performance was observed when uncertainty-based weighting was applied only during the last 20 epochs out of 100, with postprocessing disabled and deep supervision enabled. This indicates that the application of uncertainty-based weighting leads to performance degradation, while the combination of PP and DS provides the best balance among precision, recall, and IoU. Of these options, the rule-based postprocessing based on LiDAR’s physical characteristics proved simple yet highly effective. As shown in Fig.~\ref{fig:FP_compare}, the application of this step significantly reduced the amount of false-positive data.

\subsection{Datasets}
We evaluate our LIORNet using three datasets: CADC, WADS, and a self-collected snowfall dataset. The network model is trained exclusively on WADS, whereas performance evaluation is conducted across all three datasets to ensure robustness and generalization. The characteristics and roles of each dataset are detailed as follows.

First, the CADC dataset is the first adverse-weather-specific dataset designed for autonomous vehicle research. It was collected using cameras, LiDAR, and a post-processed GNSS/INS system, and includes bounding boxes for vehicles and pedestrians as well as labels for various objects such as cars, pedestrians, trucks, buses, and bicycles. However, LiDAR snow noise points are not labeled, which limits its direct use for noise removal research. Accordingly, CADC has been widely used for testing and validating the performance of noise removal algorithms \cite{pitropov2021canadian}.

Second, the WADS dataset was originally developed for autonomous driving research under snowy conditions and is widely used in LiDAR snow-noise removal studies. It contains sequential LiDAR scans with point-wise annotations that differentiate snow noise from valid environmental structures, enabling direct supervision for denoising model development. In addition, new semantic classes, such as falling snow and accumulated snow, were introduced to better capture the challenges of winter driving environments. In this study, WADS was used solely as the training dataset and served as the primary benchmark for quantitative evaluation\cite{kurup2021dsor}.

\begin{figure*}[t!]
\centerline{\includegraphics[width=0.985\textwidth]{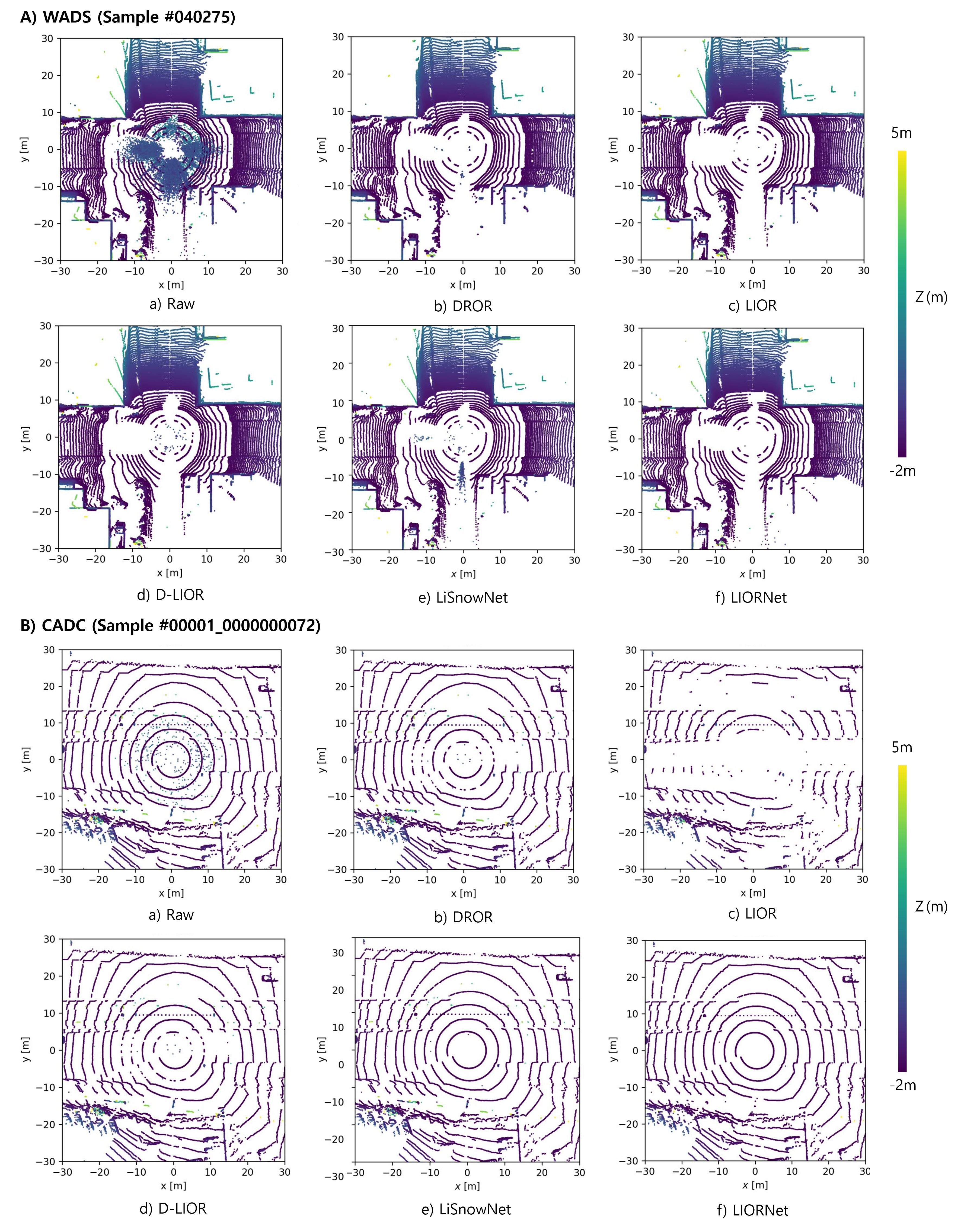}}
\caption{Comparison of performance on the (A) WADS and (B) CADC datasets. (a) Raw, (b) DROR, (c) LIOR, (d) D-LIOR, (e) LiSnowNet, (f) LIORNet.}
\label{fig:WADS_CADC}
\end{figure*}

Finally, for performance validation, self-collected snowfall data were obtained from diverse real snowfall environments in Inje, South Korea, Lule{\aa}, Sweden, and Copenhagen, Denmark. This dataset encompasses a wide range of snowfall intensities, including conditions under heavy snowfall warnings, snowfall advisories, and ordinary snow without official alerts. Such real-world data, unavailable in public benchmarks, were employed to validate the robustness and generalization capability of LIORNet.

\subsection{Filtering Performance}
\subsubsection{Baseline Methods}
To evaluate the performance of the proposed framework, we conducted comparisons with widely used representative filters and learning-based models. The baselines include LIOR~\cite{park2020fast}, which employs a fixed intensity threshold; DROR~\cite{charron2018noising}, a range-based dynamic radius method; DSOR~\cite{kurup2021dsor}, a statistical outlier removal approach; DDIOR~\cite{wang2022scalable}, a distance–intensity fusion method; D-LIOR~\cite{dlior2025}, which introduces an adaptive intensity threshold; and the recent learning-based model LiSnowNet~\cite{yu2022lisnownet}. These baselines respectively represent fixed intensity rule-based, distance/statistical, distance–intensity fusion, and learning-based paradigms, and were therefore selected as benchmarks to comprehensively assess the robustness and generalization ability of the proposed framework.

\subsubsection{Comparative Evaluation}    
Fig.~\ref{fig:WADS_CADC} presents the comparative results of various snow-removal methods on the WADS and CADC datasets. In the WADS scenario, the raw point cloud contains a large number of high-intensity noise points concentrated near the LiDAR origin, severely degrading scene visibility. The conventional DROR method removes a significant portion of the noise, but it also eliminates important environmental information such as the building wall visible in the lower part of Fig.~\ref{fig:WADS_CADC} A(b). In contrast, LIOR achieves noise reduction comparable to DROR while better preserving the surrounding environmental structures. The enhanced version, D-LIOR, improves filtering speed by updating the intensity threshold in real time using neighboring snow samples. However, as shown in Fig.~\ref{fig:WADS_CADC} A(d), when the updated threshold does not align well with the subsequent frame, it fails to properly remove snow. LiSnowNet  generally shows strong performance, but in certain frames it leaves regions where snow is not fully removed, as illustrated in the lower part of Fig.~\ref{fig:WADS_CADC} A(e). Finally, our proposed LIORNet demonstrates the most balanced results, successfully removing dense snow noise while maintaining the integrity of road boundaries and vertical structures.

On the CADC dataset, similar trends can be observed, though some notable differences are also identified. The raw input Fig.~\ref{fig:WADS_CADC} B(a) contains fewer snow points compared to WADS, but sparse snow-noise points are observed near the center. DROR Fig.~\ref{fig:WADS_CADC} B(b) alleviates part of this noise, yet its overall effectiveness remains limited. LIOR Fig.~\ref{fig:WADS_CADC} B(c) removes most of the snow, but since it relies on a fixed intensity threshold, it suffers from over-filtering in the CADC dataset with different environmental conditions, leading to the deletion of valid points. In contrast, D-LIOR Fig.~\ref{fig:WADS_CADC} B(d), although showing lower performance than LIOR on WADS, achieves more robust filtering on CADC by adopting a dynamic intensity threshold, thereby effectively suppressing snow while preserving key environmental structures such as roads and surrounding objects. LiSnowNet Fig.~\ref{fig:WADS_CADC} B(e) demonstrates better generalization than in the WADS case, producing cleaner outputs with fewer missing points. Finally, LIORNet Fig.~\ref{fig:WADS_CADC} B(f) delivers the most consistent performance across both CADC and WADS datasets, successfully eliminating the majority of snow-induced noise while maintaining important structural details.

For a fair comparison, the learning-based algorithms LiSnowNet and LIORNet were trained and tested using only the WADS dataset for up to 100 epochs. In particular, LiSnowNet was initialized with the publicly available 20 epoch pretrained weights from GitHub, and an additional 80 epochs were trained to obtain a total of 100 epoch weights for evaluation. Overall, while classical statistical filtering methods such as DROR and LIOR can provide partial noise suppression, their rule-based nature limits their ability to balance denoising and structural preservation under diverse conditions. In contrast, deep learning–based approaches, particularly LiSnowNet and our proposed LIORNet, demonstrate superior denoising performance and adaptability across different environments. Among them, LIORNet consistently achieves the best trade-off between noise suppression and the preservation of essential environmental information, thereby providing the most reliable and consistent performance across both datasets.

Table~\ref{tab:performance_compare} summarizes the quantitative evaluation results on the WADS dataset. The representative intensity-thresholding rule-based algorithm LIOR achieved higher precision, recall, and speed performance than DROR, but its processing speed was still limited to 4.8 Hz. D-LIOR, which improved this by applying a dynamic threshold, enhanced the processing speed to 8.7 Hz while maintaining overall slightly higher accuracy. However, when the dynamic threshold update is inaccurate, snow is not properly removed as shown earlier in Fig.~\ref{fig:WADS_CADC} A(d). LiSnowNet achieved very high precision (0.951) and also recorded the fastest runtime of 201 Hz, but its recall performance (0.842) was the lowest except for DROR.

In contrast, the proposed LIORNet series showed a more balanced performance. The basic U-Net based model produced unstable results, but the model applying the U-Net++ structure consistently achieved the highest recall (up to 0.961) while maintaining competitive precision (0.379–0.473). In particular, LIORNet (U-Net++ w/ PP) combined with post-processing maintained the highest recall of 0.920, while improving the precision accuracy performance—which had been relatively lower than other algorithms—up to a sufficiently competitive level, and recorded a processing speed of 43.5 Hz, thereby achieving the most outstanding overall performance balance. Although the processing speed of 43.5 Hz can be considered lower compared to LiSnowNet, considering the maximum LiDAR sensing rate of 20 Hz, it can be regarded as sufficient for real-time operation.

\begin{table*}[t!] 
\centering
\caption{Quantitative evaluation on the WADS dataset. \\ Postprocess was applied using the optimal range of 20m.}
\setlength{\tabcolsep}{12.25pt}
\renewcommand{\arraystretch}{1.25}
\footnotesize
\begin{tabular}{|c|c|c|c|c|c|c|c|}
\hline
\multirow{2}{*}{\textbf{Method}} &
\multirow{2}{*}{\textbf{Precision}} &
\multirow{2}{*}{\textbf{Recall}} &
\multirow{2}{*}{\textbf{F1-score}} &
\multirow{2}{*}{\textbf{F3-score}} &
\multirow{2}{*}{\textbf{F5-score}} &
\multicolumn{2}{c|}{\textbf{Runtime}} \\  
\cline{7-8}
 & & & & & & \textbf{ms} & \textbf{Hz} \\
\hline
DROR              & 0.764 $\pm$ 0.021  & 0.745 $\pm$ 0.152 & 0.754 & 0.747 & 0.746 & 3505 & 0.3 \\
LIOR              & 0.827 $\pm$ 0.027  & \textcolor{black}{\textbf{0.908}} $\pm$ 0.138 & 0.866 & \textbf{0.899} & \textbf{0.905} & 208.1  & 4.8  \\
D\mbox{-}LIOR     & 0.918 $\pm$ 0.038  & 0.875 $\pm$ 0.125 & \textbf{0.896} & 0.879 & 0.877 & 115.6 & 8.7 \\
LiSnowNet         & \textbf{0.951 $\pm$ 0.039} & 0.842 $\pm$ 0.167 & 0.893 & 0.852 & 0.846 & \textbf{4.98} & \textbf{201.0} \\
\hline
\textbf{LIORNet (U-Net)}    & 0.379 $\pm$ 0.107   & 0.942 $\pm$ 0.088 & 0.541 & 0.820 & 0.891 & 21.39 & 46.8 \\
\textbf{LIORNet (U-Net w/ PP)}    & 0.639 $\pm$ 0.142 & 0.922 $\pm$ 0.112 & 0.756 & 0.884 & 0.910 & 22.36 & 44.7 \\
\textbf{LIORNet (U-Net++)}  & 0.473 $\pm$ 0.136   & \textcolor{black}{\textbf{0.961 $\pm$ 0.068}} & 0.632 & 0.875 & 0.907 & 22.79 & 43.9 \\
\textbf{LIORNet (U-Net++ w/ PP)}  & \textbf{0.672 $\pm$ 0.132} & 0.920 $\pm$ 0.114 & \textbf{0.777} & \textbf{0.895} & \textbf{0.913} & \textbf{22.99} & \textbf{43.5} \\
\hline
\end{tabular}
\label{tab:performance_compare}
\end{table*}

Although numerically the lower Precision compared to Recall may suggest degraded performance, the evaluation can differ when considering the relative importance of Precision and Recall. Recall represents the accuracy of removing snow points, that is, how well the actual snow has been eliminated. Therefore, a high Recall indicates that snow points are effectively removed. In contrast, Precision denotes the proportion of non-snow environmental points that are mistakenly removed, and a low Precision means that some environmental points are eliminated. Taking LIORNet as an example, LIORNet achieves very high Recall by almost perfectly removing snow, while its Precision is relatively lower than that of the compared algorithms. However, as shown in Fig.~\ref{fig:WADS_CADC} A(f), although some environmental information is removed along with the snow, the essential environmental structures are overall well preserved, and thus there is no limitation in perceiving the surrounding environment. This demonstrates that Recall is a more important metric than Precision in this context.

This is further supported by the fact that the proportion of snow points relative to the total number of LiDAR points is very small. For instance, in a 64-channel, 1024-resolution LiDAR, approximately 65,536 points are generated per scan, of which snow points range from only a few dozen to about 10,000, corresponding to approximately 0.00076\% to 15.3\%. Considering this distribution, an F$\beta$-score analysis with an emphasis on Recall is appropriate. While F$\beta$-scores can theoretically be extended from $\beta=6$ up to $\beta=1000$, in this study we conservatively evaluated up to $\beta=5$. The results show that, for the F$5$-score, LIORNet (U-Net++ w/ PP) achieved the highest performance by maintaining a reasonable level of Precision while attaining high Recall accuracy. This suggests that, given the sparsity of snow points, LIORNet’s performance would become even more pronounced with higher $\beta$ values. These results demonstrate that LIORNet not only provides visually superior snow-noise removal performance, but also proves to be the most stable and practical method in both F$\beta$-score-based quantitative analysis and evaluations under real-world operating conditions.

\subsubsection{Ablation Study}
To investigate the contribution of each component in the proposed framework, we performed an ablation study. Thanks to the modular design of our loss function, each term can be independently enabled or disabled, allowing us to disentangle its effect on the final performance. Specifically, the analysis covers: (i) intensity-based supervision (Cond1), derived from a range-dependent intensity threshold; (ii) reflectivity-guided supervision (Cond2), based on the zero-reflectivity rule; (iii) edge-aware suppression (Cond3), restoring object edges with distance-weighted masks; (iv) the density-aware sparsity prior (Cond4), which penalizes overly dense clusters using a stop-gradient strategy; and (v) the threshold violation penalty (Cond5), which penalizes predictions that exceed intensity, range, or ground-plane constraints. In addition, we compare fixed manual weighting of loss terms against the uncertainty-based weighting scheme, in which learnable $\log\sigma$ parameters adaptively balance heterogeneous objectives.

All experiments were conducted on the WADS and CADC datasets using the LIORNet backbone. The full model includes all five loss components with uncertainty-based weighting. Variants were created by removing one component at a time (i.e., setting the corresponding weight $\lambda$ to zero) or replacing uncertainty-based weighting with fixed coefficients. This setting allows us to assess both the absolute and relative importance of each design choice. While ablation studies are typically performed by removing each term individually, in our case Cond1, Cond2, and Cond5 are sequentially dependent. Therefore, instead of excluding terms one by one, we adopt a cumulative addition strategy: starting from the baseline Intensity term (Cond1), we progressively add Reflectivity (Cond2), Edge (Cond3), Sparsity (Cond4), and finally the Penalty term (Cond5), thereby evaluating their incremental contributions.

The ablation results in Table~\ref{tab:ablation} highlight the complementary roles of all five components. The intensity-based supervision (Cond1) provides a fundamental constraint that stabilizes learning through a range–intensity relationship, and its removal causes a clear drop in Precision and IoU. Similarly, the reflectivity-guided supervision (Cond2) reduces spurious responses via the zero-reflectivity rule, and eliminating it also lowers Precision without substantially affecting Recall. The edge-aware suppression (Cond3) is the most critical factor: its removal leads to a drastic collapse in both Precision and Recall, confirming its role in preserving structural integrity at object boundaries. The sparsity prior (Cond4) further strengthens Precision and IoU by discouraging overly dense clusters, again with little influence on Recall. Interestingly, when the penalty term (Cond5) is disabled, Recall even increases slightly while Precision and IoU drop sharply. This counterintuitive effect occurs because false positives are no longer penalized, allowing more points to be classified as positives and thereby inflating Recall. In summary, Cond1–2–4 primarily enhance precision-driven robustness, Cond3 secures balanced detection performance, and Cond5 regulates the Precision–Recall trade-off by suppressing false positives at the cost of a small decrease in Recall. Together, these components complement each other, with the full model achieving the most reliable performance.

\begin{table}[t!]
\centering
\caption{Ablation study on the contributions of each component. \\
The study is conducted based on ID~4 in Table~\ref{tab:model_compare}.}
\setlength{\tabcolsep}{9.5pt}        
\renewcommand{\arraystretch}{1.35}  
\footnotesize
\begin{tabular}{|l|c|c|}
\hline
\textbf{Configuration} & \textbf{Precision} & \textbf{Recall}  \\ \hline
Baseline(Full)           & 0.672 $\pm$ 0.132 & 0.920 $\pm$ 0.114   \\   \hline
\hspace{0.2em} - Intensity term (Cond1)      & \textcolor{black}{\textbf{0.306 $\pm$ 0.094}} & 0.951 $\pm$ 0.076   \\
\hspace{0.2em} - Reflectivity term (Cond2)      & \textcolor{black}{\textbf{0.401 $\pm$ 0.110}} & 0.938 $\pm$ 0.092     \\  
\hspace{0.2em} - Edge term (Cond3)              & \textcolor{black}{\textbf{0.195 $\pm$ 0.118}} & \textcolor{black}{\textbf{0.679 $\pm$ 0.136}}  \\ 
\hspace{0.2em} - Sparsity term (Cond4)          & \textcolor{black}{\textbf{0.387 $\pm$ 0.109}} & 0.941 $\pm$ 0.090   \\ 
\hspace{0.2em} - Penalty term (Cond5)           & \textcolor{black}{\textbf{0.256 $\pm$ 0.099}} & \textcolor{black}{\underline{\textbf{0.962 $\pm$ 0.061}}}    \\ 
\hspace{0.2em} - Postprocess                    & \textcolor{black}{\textbf{0.402 $\pm$ 0.106}} & 0.939 $\pm$ 0.093   \\     
\hline
\end{tabular}
\label{tab:ablation}
\end{table}

\subsubsection{Validation on Real-world Data}
\begin{figure*}[t!]
\centerline{\includegraphics[width=1.0\textwidth]{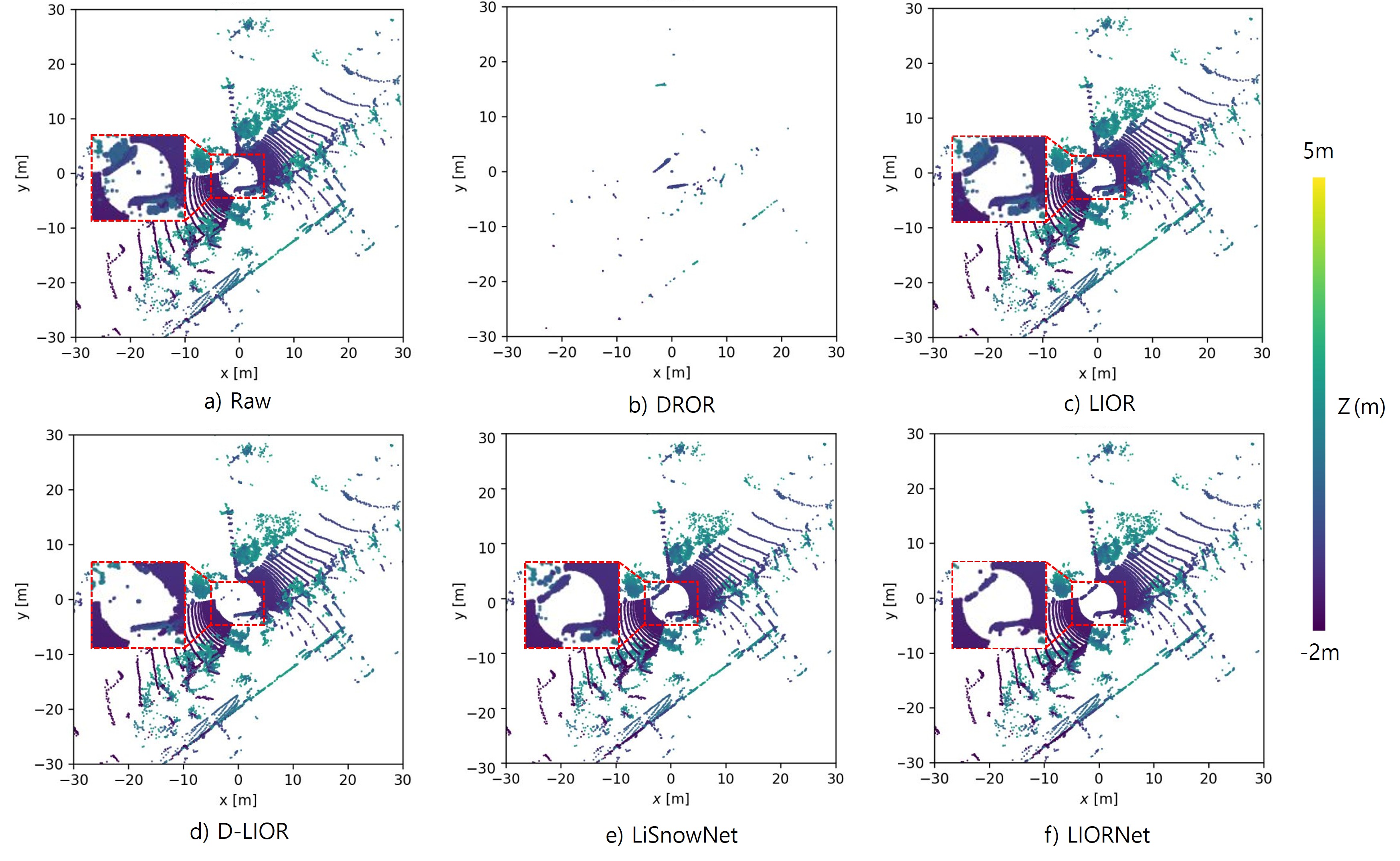}}
\caption{Performance evaluation on real-world data (a) Raw, (b) DROR, (c) LIOR, (d) D-LIOR, (e) LiSnowNet, (f) LIORNet.}
\label{fig:real_data}
\end{figure*}

Additional experiments were performed using LiDAR data acquired in real snowfall environments to validate the performance of the proposed noise-removal algorithm, and the results are presented in Fig~\ref{fig:real_data}. For a fair comparison, the same models and parameters trained on the WADS and CADC datasets were directly applied.

DROR exhibited a critical drawback in that it removed not only snow but also most of the valid points. This issue arises because the dynamic radius set in DROR does not match the LiDAR manufacturer, model, or specifications (e.g., horizontal and vertical resolution), resulting in excessive deletion of valid points. In contrast, LIOR failed to effectively remove snow, primarily because the fixed intensity threshold used in LIOR did not adapt well to differences in wavelength and signal strength across different LiDAR sensors. These results highlight that representative rule-based algorithms such as DROR and LIOR suffer from a lack of consistency when applied to different LiDAR sensors. To overcome these limitations, D-LIOR was developed, which avoided the extreme performance degradation observed in LIOR but still failed to determine an accurate intensity threshold, leaving residual snow unremoved as shown in Fig~\ref{fig:real_data}(d). These results confirm that rule-based approaches still face substantial challenges in handling diverse real-world conditions.

On the other hand, the learning-based method LiSnowNet showed relatively stable noise-removal performance in new environments compared to conventional rule-based approaches. However, some regions still exhibited a limitation in that snow was not completely removed and residual noise remained. Finally, the proposed LIORNet effectively removed dense snow while well preserving key environmental elements such as road boundaries and vertical structures, thereby providing the most balanced results. Although some major environmental elements (e.g., roadside leaves) were occasionally removed, LIORNet overall demonstrated excellent performance in accurately removing snow and clearly perceiving the surrounding environment. These results confirm that LIORNet proves its robustness and generalization capability not only on public datasets but also in real-world snowy driving environments.

\subsubsection{Summary of the Results}   
In this study, we developed LIORNet by applying distance-based, intensity-based, and learning-based approaches, and verified its noise-removal performance through comparison with representative algorithms of each approach, namely DROR, LIOR, D-LIOR, and LiSnowNet. First, our proposed pseudo-labeling method and U-Net-based self-supervised learning were proven to be effective in removing snow. By applying various network options for testing, it was confirmed that the model combining U-Net++ with deep supervision and post-processing achieved the best noise-removal performance, and the application of post-processing significantly reduced false positive snow. 

Experiments on representative snowy LiDAR datasets, WADS and CADC, showed that DROR removed a large portion of snow but also eliminated environmental information, while LIOR failed to maintain its performance from WADS when applied to CADC. To address these issues, the proposed D-LIOR showed consistent performance across both WADS and CADC, but its noise-removal capability was insufficient. Next, the learning-based algorithm LiSnowNet generally preserved environmental information while removing snow effectively, but in some WADS data with heavy snowfall, it failed to fully remove the snow. 

In contrast, our proposed LIORNet showed the most stable results by effectively removing dense snow while preserving key elements such as road boundaries and vertical structures not only in WADS and CADC but also in real-world acquired data. 
(Although its inference speed is somewhat lower than that of the comparison algorithms in terms of accuracy–speed trade-off, considering the maximum LiDAR sensing rate of 20 Hz, the achieved 20 FPS indicates that real-time noise removal is feasible and sufficient for practical autonomous driving applications.)
Overall, LIORNet, by integrating all approaches, proved to be the most effective and generalizable method, combining the advantages of existing methods, overcoming the lack of consistency in rule-based approaches, and achieving superior accuracy in noise removal compared to existing learning-based methods. In particular, its significance lies in demonstrating the effectiveness through training with pseudo-labeled data and employing the image segmentation network U-Net.

\section{Conclusion}
In this paper, we presented LIORNet, a self-supervised LiDAR snow-removal framework that integrates distance-based, intensity-based, and learning-based approaches within a unified U-Net++ architecture. The key idea lies in the use of physics- and statistics-guided pseudo-labels, generated through five complementary conditions (Cond1–Cond5: range-dependent intensity, zero-reflectivity rule, edge-aware suppression, density-aware sparsity, and threshold violation penalty). This modular design enables the network to effectively learn snow suppression without manual annotation, while uncertainty-based loss weighting adaptively balances heterogeneous objectives.

The experimental results demonstrated that LIORNet outperformed conventional rule-based filters (DROR, DSOR, DIOR/DDIOR, LIOR/D-LIOR) and the state-of-the-art learning-based method LiSnowNet in terms of precision, while its recall was relatively lower. Nevertheless, when considering the relative importance through the F-score, LIORNet achieved overall superior performance. Furthermore, despite the relatively lower recall, it was shown that critical environmental information essential for autonomous driving, such as road boundaries and vertical structures, was sufficiently preserved. Ablation studies further confirm that each loss term contributes positively, and that sequentially adding Cond2–Cond5 on top of the baseline Cond1 progressively enhances performance, validating both the modularity and the uncertainty-based weighting strategy. Nevertheless, LIORNet imposes higher computational overhead than traditional rule-based filters, which may hinder its direct application in embedded or real-time systems. Furthermore, its reliance on pseudo-label quality makes it sensitive to extreme snowfall, rare boundary conditions, and cross-sensor variability.

For future research, we plan to extend LIORNet by directly leveraging raw 3D point clouds instead of 2D projections, incorporating temporal information to improve both efficiency and accuracy, and generalizing beyond snow to other adverse weather conditions such as rain and fog. This extended real-time framework, RT-LIORNet, is expected to advance robust perception and contribute toward achieving fully autonomous driving at Level 5 under all environmental conditions.

\section*{Acknowledgments}
This should be a simple paragraph before the References to thank those individuals and institutions who have supported your work on this article.

\bibliographystyle{IEEEtran}
\bibliography{reference}

\vspace{11pt}

\begin{IEEEbiography}
[{\includegraphics[width=1in,height=1.25in,clip,keepaspectratio]{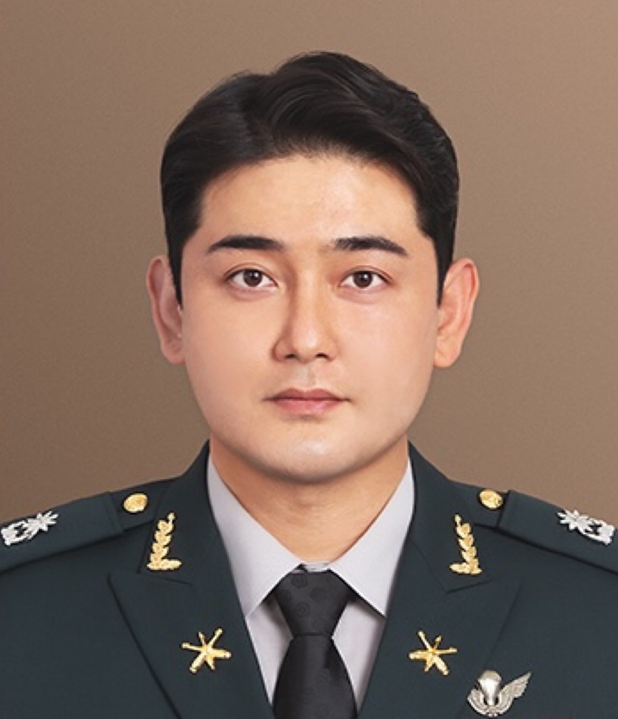}}]{Ji-il Park} received the B.S. degree in mechanical engineering from Korea Military Academy, Seoul, South Korea, in 2005, and the M.E. and Ph.D. degrees in mechanical engineering from KAIST, Daejeon, South Korea, in 2013 and 2022, respectively. He served as an Assistant Professor with the Department of Mechanical Engineering, Korea Military Academy, until 2019. He served as an Officer in charge of robot and AI at Defense Innovation Bureau, Office of the Vice Minister of National Defense, Ministry of National Defense, from 2022 to 2023, and served at the Defense AI Center in the Agency for Defense Development from 2024 to 2025, where he was responsible for artificial intelligence policy and the requirements planning of ground AI weapon systems. He is currently a Lieutenant Colonel with the Ministry of National Defense and holds a concurrent position at the Department of Smart Mobility Engineering, Inha University.
\end{IEEEbiography}

\begin{IEEEbiography}
[{\includegraphics[width=1in,height=1.25in,clip,keepaspectratio]{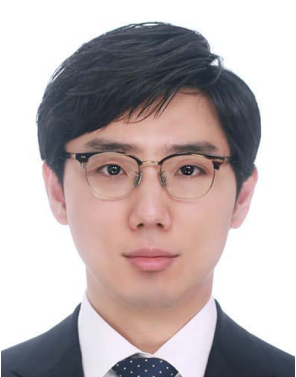}}]{Inwook Shim} received the B.S. degree in computer science from Hanyang University, in 2009, the M.S. degree in Robotics Program in electrical engineering from KAIST,
in 2011, and the Ph.D. degree in electrical engineering from the Division of Future Vehicles, KAIST, in 2017. From 2017 to 2022, he worked as a Research Scientist at Agency for Defense Development, South Korea. He is currently an Assistant
Professor with the Department of Smart Mobility Engineering, Inha University, Incheon, South Korea. His research interests include 3D vision for autonomous systems and deep learning. He was a member of Team KAIST, which took first place at the DARPA Robotics Challenge Finals, in 2015. He was a recipient of the Qualcomm Innovation Award, and a NI finalist at the NI Student Design Showcase. He received the KAIST Achievement Award of Robotics, and the Creativity and Challenge Award from KAIST.
\end{IEEEbiography}  

\vfill

\end{document}